\documentclass[conference]{IEEEtran}
\usepackage{cite}
\usepackage{amsmath,amssymb,amsfonts}
\usepackage{algorithmic}
\usepackage{graphicx}
\usepackage{textcomp}
\usepackage{xcolor}
\graphicspath{{/}}
\def\BibTeX{{\rm B\kern-.05em{\sc i\kern-.025em b}\kern-.08em
    T\kern-.1667em\lower.7ex\hbox{E}\kern-.125emX}}
\IEEEoverridecommandlockouts
\begin{document}

\title{A Novel Graphic Bending Transformation on Benchmark\\
}

\author{\IEEEauthorblockN{Chunxiuzi Liu$^{1}$, Fengyang Sun$^{1}$, Qingrui Ni$^{1}$, Lin Wang$^{1,*}$, Bo Yang$^{1,2}$}
\IEEEauthorblockA{\textit{$^{1}$Shandong Provincial Key Laboratory of Network based Intelligent Computing, University of Jinan, Jinan, China} \\
\textit{$^{2}$School of Informatics, Linyi University, Linyi, China}\\
 $^{*}$Corresponding Author, Email: wangplanet@gmail.com\\
}
\thanks{C. Liu and F. Sun——Contribute equally to this article. }
}

\maketitle

\begin{abstract}
Classical benchmark problems utilize multiple transformation techniques to increase optimization difficulty, e.g., shift for anti centering effect and rotation for anti dimension sensitivity. Despite testing the transformation invariance, however, such operations do not really change the landscape's ``shape'', but rather than change the ``view point''. For instance, after rotated, ill conditional problems are turned around in terms of orientation but still keep proportional components, which, to some extent, does not create much obstacle in optimization. In this paper, inspired from image processing, we investigate a novel graphic conformal mapping transformation on benchmark problems to deform the function shape. The bending operation does not alter the function basic properties, e.g., a unimodal function can almost maintain its unimodality after bent, but can modify the shape of interested area in the search space. Experiments indicate the same optimizer spends more search budget and encounter more failures on the conformal bent functions than the rotated version. Several parameters of the proposed function are also analyzed to reveal performance sensitivity of the evolutionary algorithms.
\end{abstract}


\section{Introduction and Motivation}
\begin{figure}[tb]
    \centering
    \includegraphics[width=0.3\textwidth]{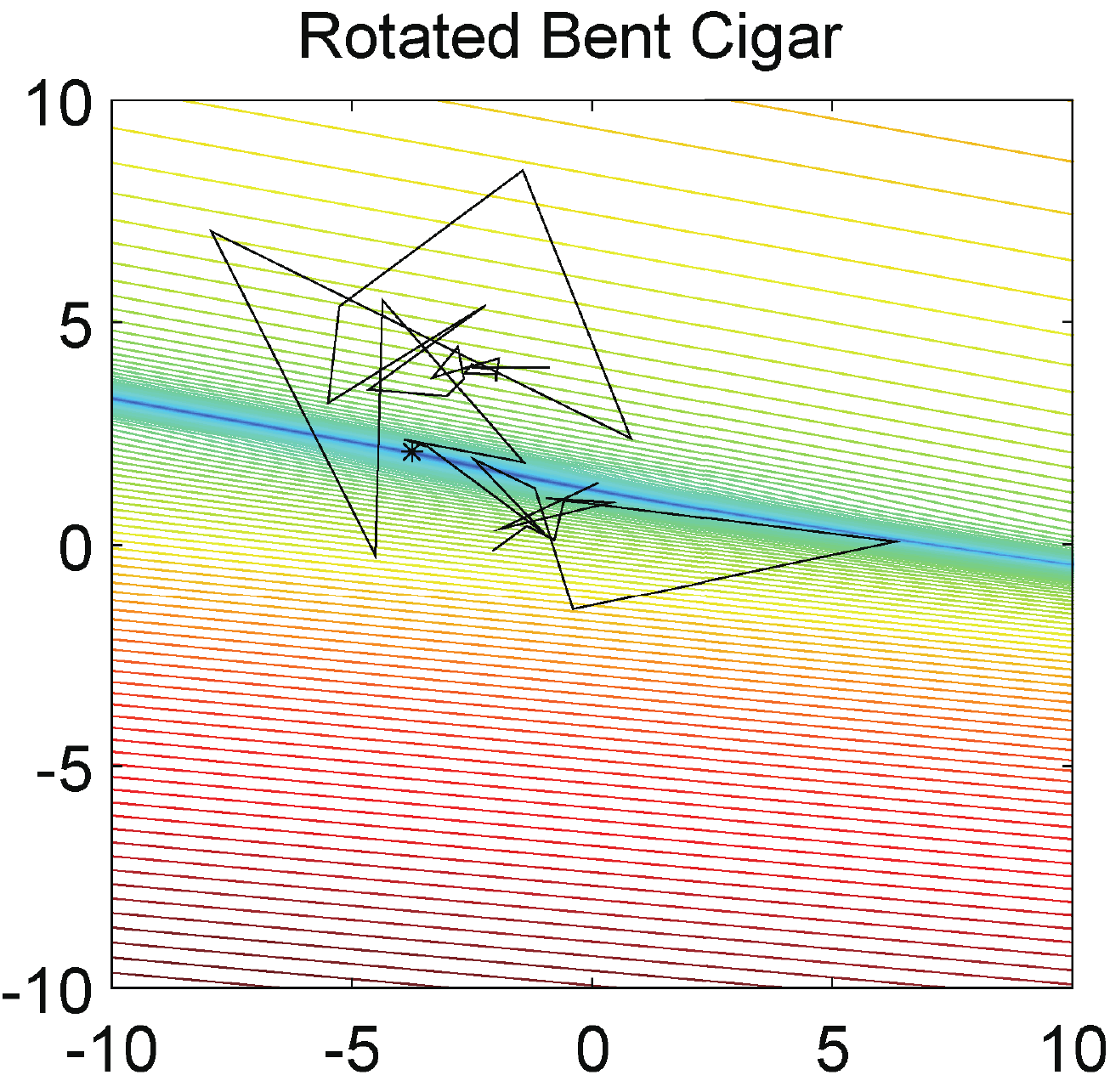}
    \caption{Visualization for search trace of CMA-ES on a rotated ill conditional function (the bent cigar function). The trace is drawn with black line and the optimum is denoted with black $*$.
    }
    \label{fig-spark}
\end{figure}

The ill conditional property of an optimization problem, generally concerns that the sensitivity of function value changes in terms of different variables or search directions is distinct \cite{Hansen2011Impacts}. In other word, an ill conditional function is difficult to be optimized because it has such properties: the change in some variables or directions does slight influence on function value, but a minute change in other variables or directions can instill a drastically substantial change in function value \cite{Saarinen1993illConditioningNN,Hariya2016PSOnonSepillCond}. The previous literature has empirically shown that covariance matrix adaptation evolution strategy (CMA-ES) is efficient in solving such problems \cite{Hansen2008PSOnonSepillCond}. We can observe from Fig. \ref{fig-spark} that the optimizer spent some time on exploring the best region over the general  search space (most areas of which are undesirable) in the early stage. However, once it positioned the best region, which is the ``banded long narrow valley'', the superior local search capability of CMA-ES can guide the search agent to rapidly approach the optimum with negligible demand of further tuning search directions \cite{Hansen2008PSOnonSepillCond,2017Stability,2020Stability}.

\par
The intuitive inference could be that, the ``straight banded valley'' provides some convenience for optimization. Since the principal search direction of this area is a unique straight line, a strategy similar to line search can gain substantial benefits after it decides the right search direction around the area\cite{2017Optimal,2018Dendritic,2017Suboptimal}. Therefore, the question arises here is, \emph{what if the valley is winding instead of straight?}

\par
Classical benchmark constructing approaches test optimizer generalization and invariance in some aspects by orthogonal transformations such as rotation. Nevertheless, such transformations does not really change the function ``shape'', but rather than achieve certain adjustments in the sense of ``view point''. For instance, in two dimensional scenario, an ill conditional problem simply changes the eye focus horizontally or vertically after translated, and changes the eye focus rotationally after rotated (see Fig. \ref{fig-VisualEffectRawRotConformal}). However, these transformations both keep the straight line shape of raw function. By contrast, the conformal mapping in topology maintain some basic properties such as local angles and not necessarily maintain others such as lengths when processing images \cite{Schinzinger2012Conformal}. These characteristics enable conformal transformation altering the general appearance of an ill conditional function but basically still keep its unimodality \cite{Sharon20062dConformal}.

In this paper, we introduce a novel bending function operation with conformal mapping to test the search behavior changes of evolutionary algorithms. We construct a conformal ill conditional function in two dimensional case. The CMA-ES  and particle swarm optimization (PSO) \cite{eberhart1995new} algorithms are evaluated and considerable performance changes are observed based on the behavior statistics. In addition, we demonstrate the involved optimizers are sensitive to positions of the optimum and starting search points by tuning the problem parameters.

\section{Conformal Mapping}

\subsection{General Description}
\par
Conformal mapping is a branch of complex variable function theory.
It studies complex variable functions from a geometric point of view, which maps a region to another region through an analytic function.
Conformal transformation maintains the angle and the shape of infinitesimal objects.
Unlike the traditional translation mapping or rotation, it does not necessarily maintain their size.
Conformal mapping has two important properties: rotation angle invariance and scaling invariance.
The reader needs to distinct them from the similar descriptions mentioned in \cite{Hansen2011Impacts}.

\par
Let the function $f(z)$ be parsed in the region $\Omega$, $z_0 \in \Omega$, and $f{(z_{0})}' \neq 0$.
Rotation angle invariance means that the angle between two curves passing a point $z_0$ in the area $\Omega$ and the angle between the two curves obtained after mapping remain unchanged in size and direction.
Scaling invariance means that the scaling rate of any curve passing through $z_0$ is $\left | f{(z_{0})}' \right | $, regardless of its shape and direction.
If the above conditions are met, we call $ w=f(z_{0}) $ a conformal mapping at $z_0$ \cite{Schinzinger2012Conformal}.

\begin{figure}[tb]
    \centering
    \includegraphics[width=0.35\textwidth]{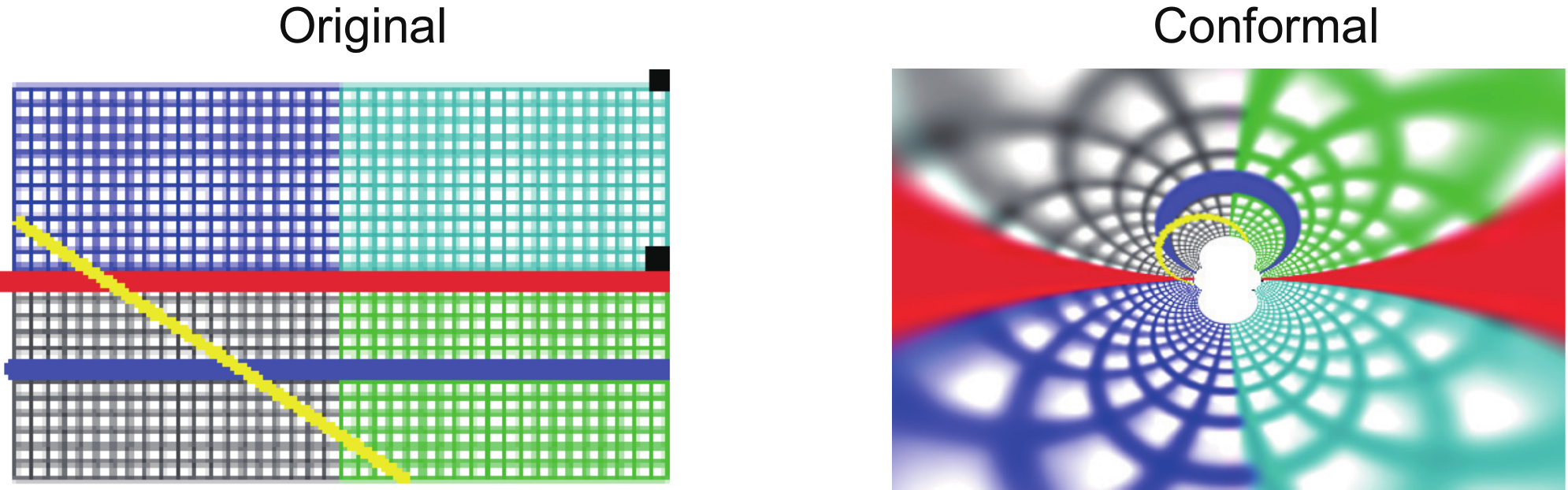}
    \caption{A grid image is transformed by conformal mapping. The blue horizontal line and the yellow line have been bent as circles.
    }
    \label{fig-imageConformalEffect}
\end{figure}

\subsection{Linear Fractional Transformation}
\par
Linear fractional transformation is a commonly used mapping, the basic form of which is $w= f(z)= \frac{az+b}{cz+d}$, where $z$ and $w$ are complex variables and $a,b,c,d,\in \mathbb{R}$.
Let $w$ take a derivative w.r.t. $z$, and the outcome is $ \frac{\partial w}{\partial z}=\frac{ad-bc}{\left ( cz+d \right )^{2}}$.
According to the definition of conformal mapping, we can draw a conclusion that $f(z)$ is a form of conformal mapping if $ad-bc\neq 0$.

\begin{equation}\label{eq-linear fractional transformation}
w = \frac{{az + b}}{{cz + d}} = \left\{ {\begin{array}{*{20}c}
   {\frac{a}{d}(z + \frac{b}{a})}, {c = 0}  \\
   {\frac{a}{c} + \frac{{bc - ad}}{{c(cz + d)}}}, {c \ne 0}  \\
\end{array}} \right.
\end{equation}

\par
As shown in Eq. \ref{eq-linear fractional transformation}, any fractional linear map can be decomposed into three basic maps: translation mapping $w=z+b$, similarity mapping $w=az$, and inversion mapping $w=\frac{1}{z}$.
Linear fractional transformation has the properties of one-to-one correspondence and circularity-preserving on the extended complex plane, that is, the circle before the mapping (a straight line is regarded as a circle passing through infinity) is still a circle after mapping.

\par
We use the inversion mapping to process the image, and find that the inversion transform distorts the image content (Fig. \ref{fig-imageConformalEffect}), which is difficult to be implemented by traditional orthogonal transformations such as translation or rotation.
In this way, we can infer that if the search path is inversely transformed, the distortion degree of the path can be changed, resulting in the problem with varying search difficulty.

\begin{figure*}[ht]
    \centering
    \includegraphics[scale=0.35]{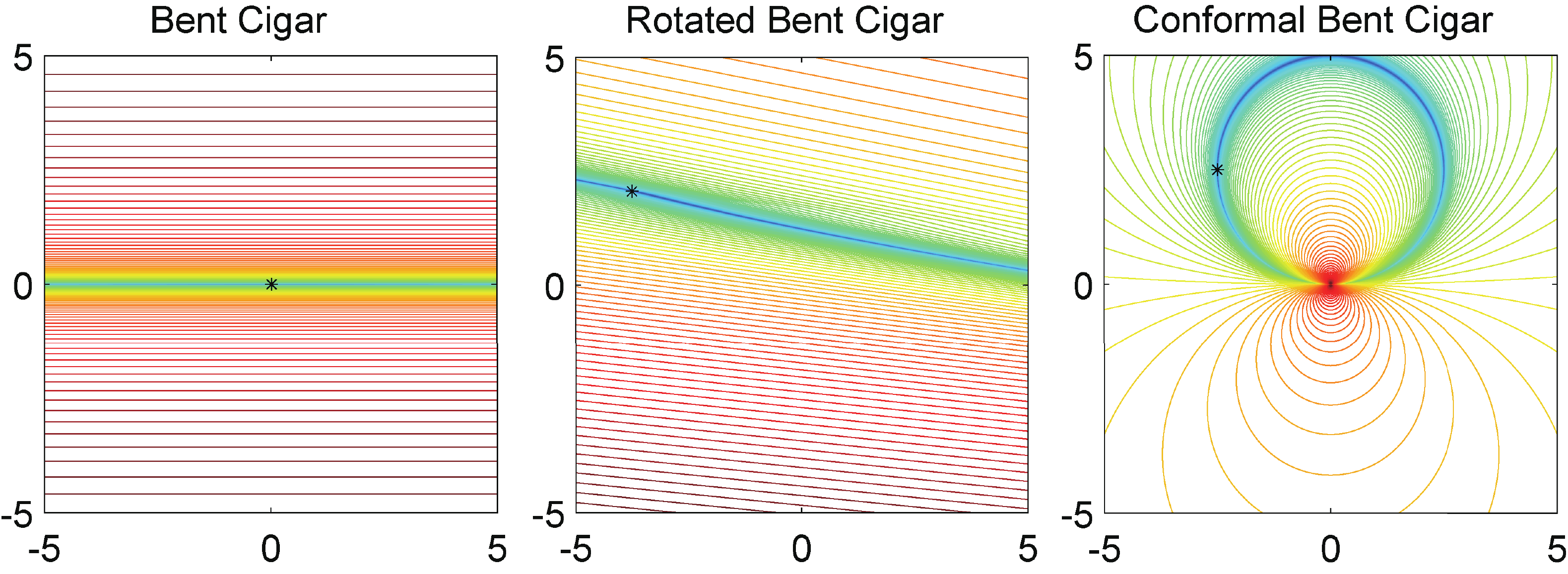}
    \caption{Visualization for different versions of the bent cigar function. The black $*$ denotes position of the optimum.
    }
    \label{fig-VisualEffectRawRotConformal}
\end{figure*}

\section{Constructing Conformal Landscapes}
\par
The conformal version of an ill conditional function is constructed by mapping the input position vector with $w = 1/z$ before computing the function value. The entire mapping process consists of three steps:
\begin{enumerate}
\item forward box transformation to shrink the original search space
\item conformal mapping to change the function shape
\item inverse box transformation to recover the search space
\end{enumerate}
The input position vector is generally formed with ${\bf{x}} = (x_1 ,x_2 , \ldots ,x_D )$, where $D$ is the problem dimensionality. In this paper, we specifically focus on two dimensional problem, which makes ${\bf{x}} = (x_1 ,x_2 )$. The following contents of this section explain the implementation details of the three steps.
\subsection{Forward Box Transformation}

\par
We use the constructed scale factor ${\bf{s}}_{{\rm{forward}}}$ and shift factor ${\bf{o}}_{{\rm{forward}}}$ to scale and translate the input vector, respectively. For each input vector ${\bf{x}} \in \Omega$, after forward box transformation, we get a new form of $x^\prime$ in the domain $\Omega^\prime$. The domain $\Omega^\prime$ is usually much smaller than the original domain $\Omega$. The computation of ${\bf{x}}^\prime$ are shown in Eqs. \ref{eq-sForward}, \ref{eq-oForward}, and \ref{eq-forwardBoxTransf}.

\begin{equation}\label{eq-sForward}
{\bf{s}}_{{\rm{forward}}}  = (\frac{{2\xi }}{L}, - \frac{{2\psi }}{L}),
\end{equation}
\begin{equation}\label{eq-oForward}
{\bf{o}}_{{\rm{forward}}}  = (- \xi ,\psi ),
\end{equation}
\begin{equation}\label{eq-forwardBoxTransf}
{\bf{x}}^\prime = {\bf{x}} \circ {\bf{s}}_{{\rm{forward}}}  + {\bf{o}}_{{\rm{forward}}} ,
\end{equation}

\noindent where $\xi$ and $\psi$ are 2 hyper parameters controlling function shape after deformed. $L$ denotes the scale level of the space, empirically set equal to the length of each dimension of the hypercube.In order to facilitate the calculation of the best position, we set ${\bf{o}}_{{\rm{forward}}}  = (0,0)$ for simplicity in this paper.

\subsection{Conformal Transformation}

\par
At this step, we put the vector ${\bf{x}}^\prime$ in the complex plane and conduct conformal inversion mapping to obtain the mapped vector ${\bf{x}}^{\prime \prime }$. The computation of ${\bf{x}}^{\prime \prime }$ are shown in Eqs. \ref{eq-zComplex}, \ref{eq-wConformalInv}, and \ref{eq-xConformal}.

\begin{equation}\label{eq-zComplex}
z = {\rm{complex}}({\bf{x'}}) = x_1 ^\prime   + x_2 ^\prime  {\rm{i}},
\end{equation}
\begin{equation}\label{eq-wConformalInv}
w = \frac{1}{z} = \frac{1}{{x_1 ^\prime   + x_2 ^\prime  {\rm{i}}}} = \frac{{x_1 ^\prime  }}{{x_1 ^\prime { ^2}  + x_2 ^\prime  {^2 }}} - \frac{{x_2 ^\prime  }}{{x_1 ^\prime  {^2}  + x_2 ^\prime  {^2} }}{\rm{i}},
\end{equation}
\begin{equation}\label{eq-xConformal}
{\bf{x}}^{\prime \prime } = {\rm{decomplex}}(w) = (\frac{{x_1 ^\prime  }}{{x_1 ^\prime  {^2 } + x_2 ^\prime { ^2 }}}, - \frac{{x_2 ^\prime  }}{{x_1 ^\prime { ^2}  + x_2 ^\prime  {^2} }}),
\end{equation}

\noindent where the function complex transforms a two dimensional vector into a complex number and function decomplex transforms a complex number into a two dimensional vector.

\begin{figure*}[h]
    \centering
    \includegraphics[scale=0.36]{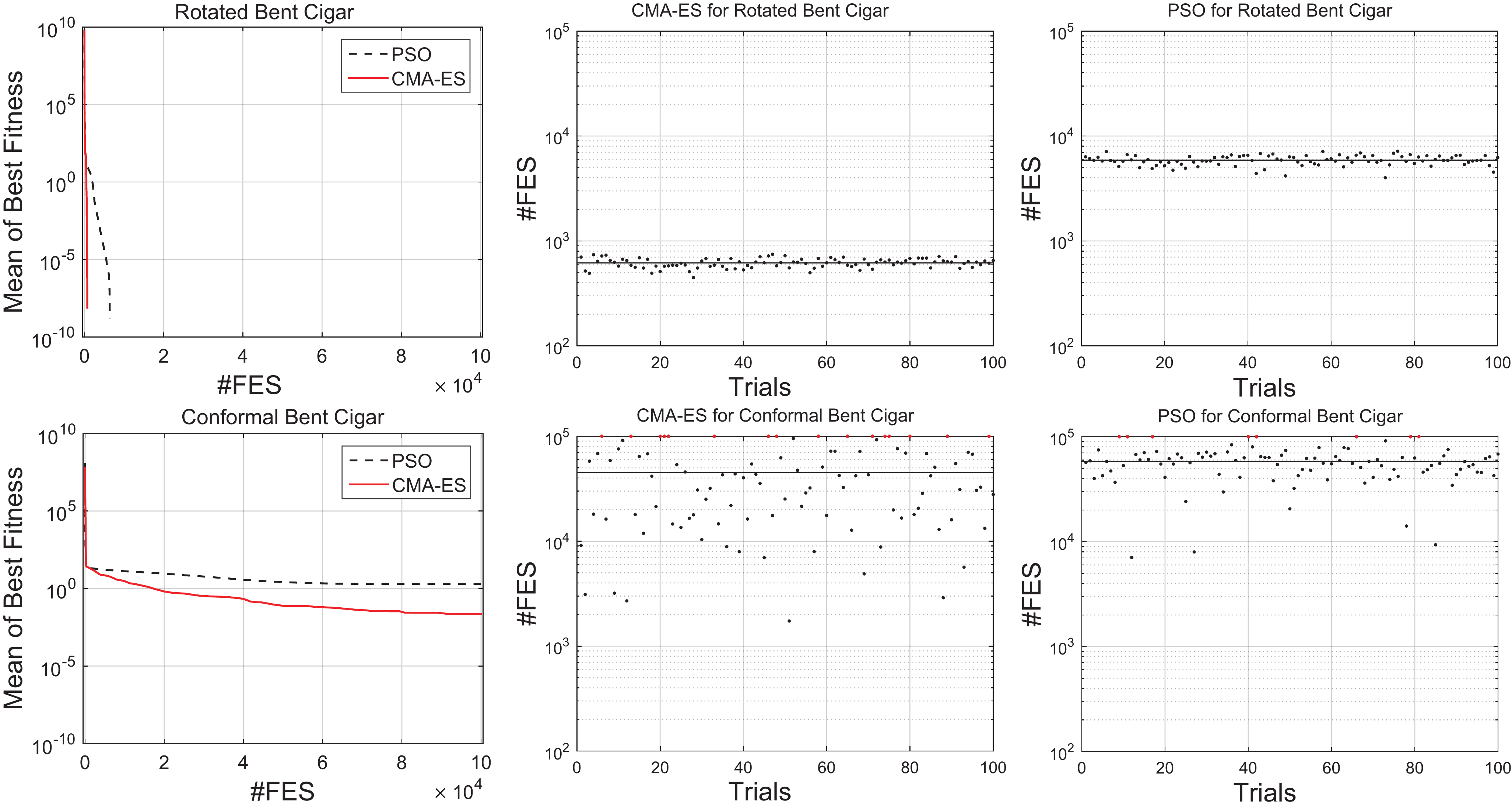}
    \caption{The convergence distinction and $\#FES$ distribution for CMA-ES and PSO on both rotated and conformal bent cigar functions over 100 independent repeated trials. The black horizontal solid line in the right subfigure represents average number of fitness evaluations used to find the optimum. The black dots represents $\#FES$ used for each trial and the red dots for the records that the algorithm failed. The smaller $\#FES$ and lower fitness value indicate better performance.
    }
    \label{fig-ConvergeScatterRotConfmAll}
\end{figure*}

\subsection{Inverse Box Transformation}

\par
After conformal mapping, we get the new vector ${\bf{x}}^{\prime \prime }$ in the domain $\Omega^\prime$. In order to get the new form of ${\bf{x}}$ mapped in the original domain $\Omega$, we use the constructed inverse scaling factor ${\bf{s}}_{{\rm{inverse}}}$ and the inverse translation factor ${\bf{o}}_{{\rm{inverse}}}$ to transfer it inversely to $\Omega$, noted ${\bf{x}}^{\prime \prime \prime}$. Then ${\bf{x}}^{\prime \prime \prime}$ can be used to calculate the real function value. The computation of ${\bf{x}}^{\prime \prime \prime}$ are shown in Eqs. \ref{eq-sInverse}, \ref{eq-oInverse}, and \ref{eq-inverseBoxTransf}.

\begin{equation}\label{eq-sInverse}
{\bf{s}}_{{\rm{inverse}}}  = (\frac{{2\upsilon }}{L}, - \frac{{2\varpi }}{L}),
\end{equation}
\begin{equation}\label{eq-oInverse}
{\bf{o}}_{{\rm{inverse}}}  = (- \upsilon ,\varpi ),
\end{equation}
\begin{equation}\label{eq-inverseBoxTransf}
{\bf{x}}^{\prime \prime \prime}= ({{{\bf{x}}^{\prime \prime} - {\bf{o}}_{{\rm{inverse}}} }})\oslash{{{\bf{s}}_{{\rm{inverse}}} }},
\end{equation}
\noindent where $\upsilon$ and $\varpi$ are 2 hyper parameters controlling size of the ``valley'' and position of the optimum after deformed. For avoiding any confusion, we here use $\circ$ and $\oslash$ to represent Hadamard element-wise multiplication and division of vectors, respectively.

Fig. \ref{fig-VisualEffectRawRotConformal} demonstrates the visual differences among the raw, rotated and conformal bent cigar functions. The function is bent as a ring shape by conformal mapping as expected. In this scenario, the optimizer could be forced to carefully adjust the direction all the time in the winding path and present poor efficiency.

\section{Experiments}

\par
In this section, we explore how the winding shape of the ill conditional function impacts search.
The results show that the optimizers are extremely hard to find the optimum after the ill conditional function is bent by conformal mapping.
In addition, narrower valley around the optimum contributes more difficulty. 

\par
Since multi restart CMA-ES \cite{Hansen2001CMAES} is empirically regarded efficient on ill conditional problems and PSO \cite{ElAbd2009PSO} on the contrary, we here focus on them to test the performance variation with respect to function shape \cite{Hansen2011Impacts}. The reader is referred to the corresponding citations for the detailed parameter settings and main source codes of these optimization methods.

\par
The ill conditional function \emph{bent cigar} \cite{Hansen2009BBOBExpsetup} is used as the basic function to be further transformed. The default parameter configurations are $\xi$, $\psi$, $\upsilon$, $\varpi$ = 1, $L=10$.
We choose the expected running time (ERT) \cite{Hansen2009BBOBExpsetup} and mean best fitness to measure the performances of the involved algorithms. The calculation formula of the ERT is shown in Eq. \ref{eq-ERT}.

\begin{equation} \label{eq-ERT}
ERT = RT_s  + \frac{{1 - p_s }}{{p_s }}RT_{us} ,
\end{equation}

\noindent where $RT_s$ and $RT_{us}$ denote the average number of function evaluations for successful and unsuccessful runs, respectively. $p_s$ is the fraction of successful runs. ``Successful'' means that the algorithm reaches the predefined accuracy level ($10^{-6}$) on the test function, which, to some extent, corresponds to the best solution. $\#FES$ denotes the number of function evaluations, and maximum of $\#FES$ is set as $10^5$. Unless specified, the configurations of all the experiments remain the same.
\begin{figure}[tb]
    \centering
    \includegraphics[width=0.45\textwidth]{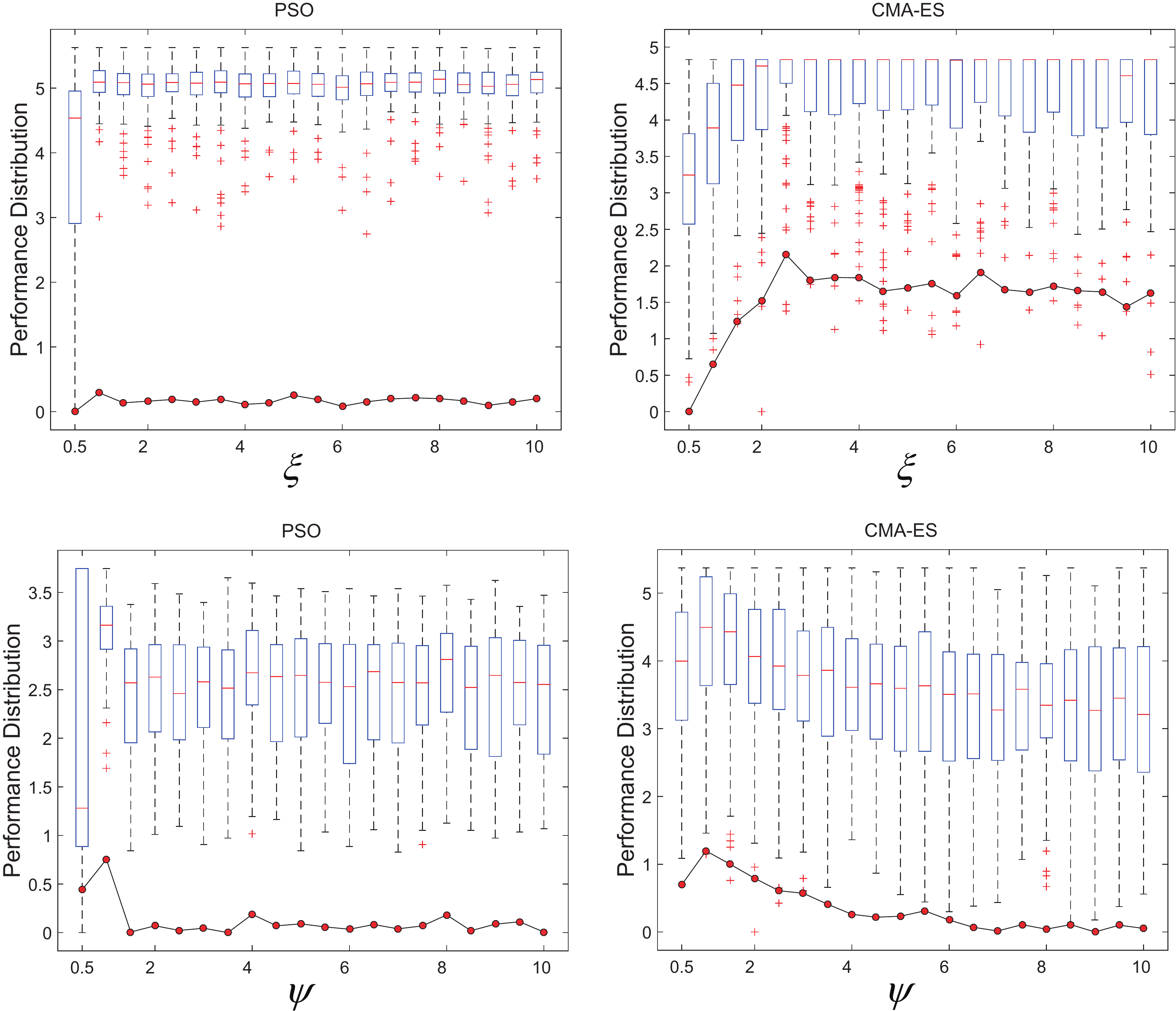}
    \caption{The $\#FES$ and ERT over 100 trials versus increasing hyper parameters $\xi$ and $\psi$.
    The other parameters keep unchanged when testing one parameter.
    The raw values are divided by the minimum value and logarithmically scaled to enhance the contrast.
    }
    \label{fig-boxplotERTxdyd}
\end{figure}
\begin{figure}[tb]
    \centering
    \includegraphics[width=0.45\textwidth]{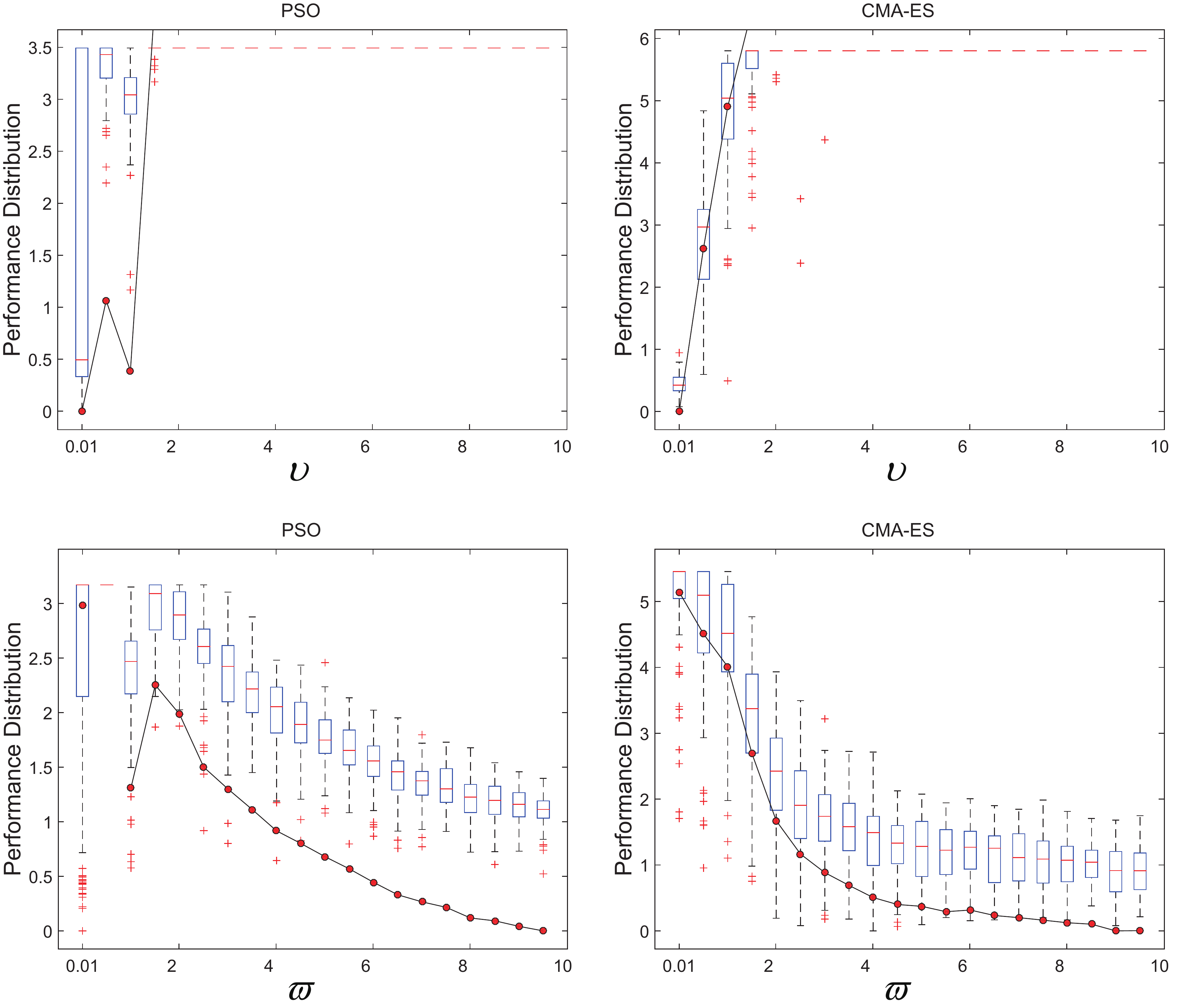}
    \caption{The $\#FES$ and ERT over 100 trials versus increasing hyper parameters $\upsilon$ and $\varpi$.
    The other parameters keep unchanged when testing one parameter.
    The raw values are divided by the minimum value and logarithmically scaled to enhance the contrast.
    Part of the ERT curves are not shown since the corresponding ERT values are infinite, which indicates all of the trials under the same parameter value failed.
    }
    \label{fig-boxplotERTudvd}
\end{figure}

\subsection{Comparison of Ring and Line Shape on Convergence}

\par
Fig. \ref{fig-ConvergeScatterRotConfmAll} illustrates the distinct convergence behavior of the involved algorithms between rotated and conformal ill conditional functions.
The rotated version of bent cigar function is relatively easy to be optimized for both CMA-ES and PSO.
However, the difficulty is dramatically increasing after the function is transformed by conform mapping.
The fitness values are continuously decreasing across the whole search process but in an unbearable slow speed, especially in the late phase of optimization. The larger performance variations also reveal the instability and sensitivity of the optimizers on the conform problem, since the only difference over various trials is the starting search point.

\par
Albeit CMA-ES outperformed PSO in terms of mean fitness values on both problems, the performance gap was narrowed on the conform one. Despite more trials that find the optimum with smaller $\#FES$, CMA-ES presented larger performance variation and two times more failed trials than PSO.

\subsection{Impact of Ring Shape}

\par
\begin{figure*}[h]
    \centering
    \includegraphics[scale=0.30]{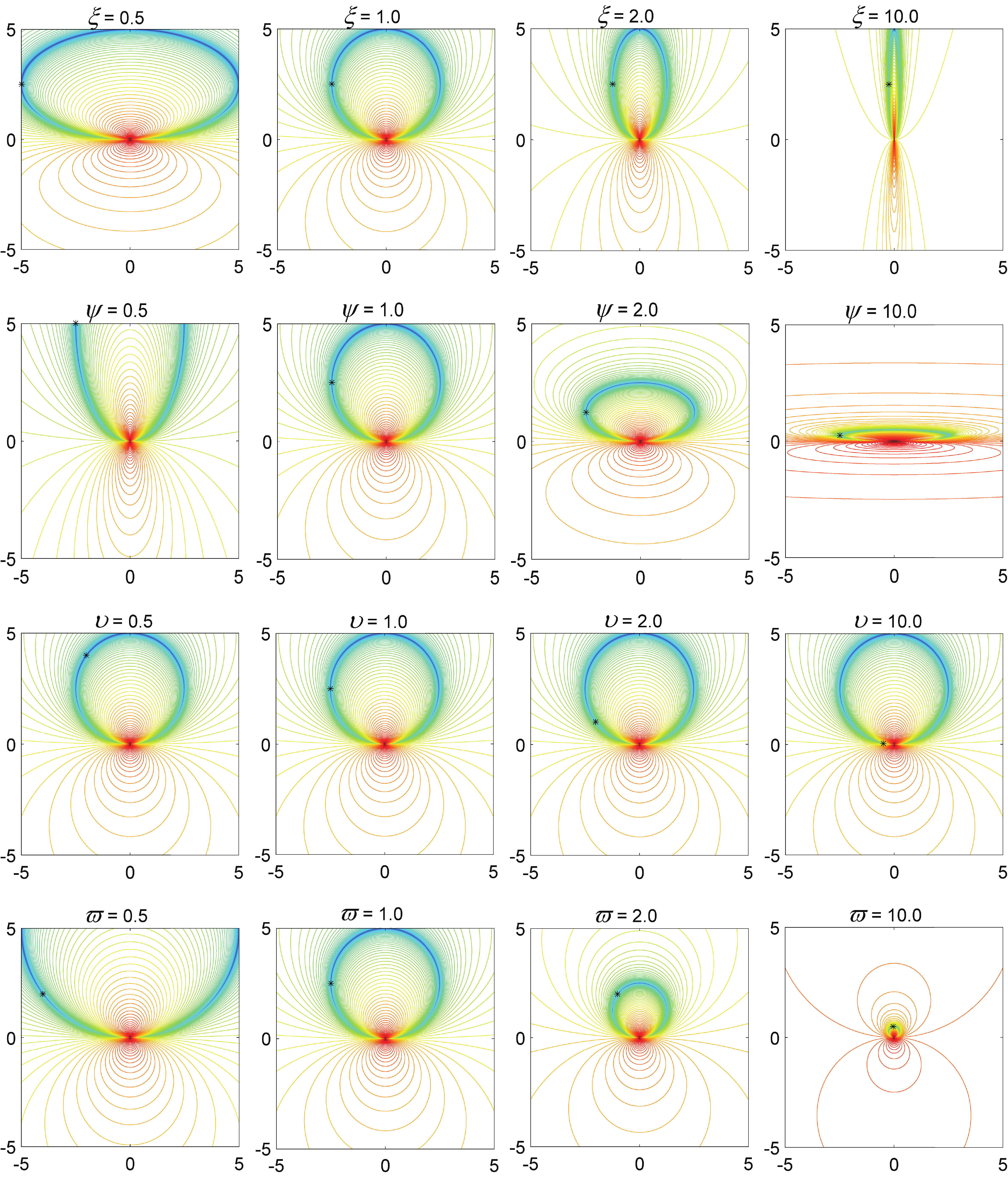}
    \caption{Contour maps for visualizing two dimensional conform bent cigar functions with various hyper parameters.
    The black $*$ denotes position of the optimum.
    When one parameter is changing, the others keep fixed, which are the default configurations $\xi$, $\psi$, $\upsilon$, $\varpi$ = 1.
    }
    \label{fig-visualContourParams}
\end{figure*}

The parameter analysis of the conform operation are summarized in numerical results (Fig. \ref{fig-boxplotERTxdyd} and \ref{fig-boxplotERTudvd}) and visualizations (Fig. \ref{fig-visualContourParams}).
From Fig. \ref{fig-visualContourParams} top two rows, the parameters $\xi$ and $\psi$ primarily impact the ring shape by tuning the proportion between the ``long axis'' and ``short axis'' of the ring. As $\xi$ increases, the ring becomes lanky. Although the curvature of the search direction around the optimum is getting smaller, the width of the ring is also narrowing with increasing $\xi$. As shown in Fig. \ref{fig-boxplotERTxdyd}, this case does not bother PSO much, but produce considerable difficulty for CMA-ES.

\par
As $\psi$ increases, the ring becomes prolate. Since the width of the ring is not narrowing severely compared to $\xi$, and the ring area far from the optimum is also becoming ``soft winding'' (smaller curvature), the search agent is able to approach the optimum along the ring. Therefore, CMA-ES was less influenced by $\psi$ compared to $\xi$ and performed better as $\psi$ increases (Fig. \ref{fig-boxplotERTxdyd}).

\subsection{Impact of Ring Size and Optimum Position}
\par
From Fig. \ref{fig-visualContourParams} bottom two rows, the parameters $\upsilon$ and $\varpi$ primarily impact the ring size and optimum position of the conformal bent cigar function. As $\upsilon$ increases, the basic function shape is not changing but the optimum position is being moved to the ``thinner'' area, which is closed to highly undesirable red points. In this case, the search agent is easier to make mistakes around the subtle ring dot like walking high wire. The CMA-ES and PSO both present terrible performance under larger $\upsilon$ value, as shown in Fig. \ref{fig-boxplotERTudvd}. When the value of $\upsilon$ is larger than 2, the involved optimizers can not find position the optimum in any trial within preset number of fitness evaluations.

\par
As $\varpi$ is increasing, the ring size is becoming smaller. Although the width of the ring is also narrowing, the entire area of best region is shrinking due to the contracted perimeter, which shortens the time finding the optimum, regardless where the starting point is. Fig. \ref{fig-boxplotERTudvd} has validated the explanation. As $\varpi$ increases, CMA-ES and PSO both present better performance.

\section{Conclusion}

\par
The shape of local valley in the ill conditional landscape can clearly impact the optimizer performance.
After bent using conformal transformation, it creates substantial barrier in the search process for both CMA-ES and PSO algorithms.
The optimizers spend 100 times more search budget on solving it and even still fail to position the optimum for many trials, which is a quite rare case in the optimization of the rotated version of bent cigar function. CMA-ES presents higher sensitivity on valley shape change caused by hyper parameters than PSO.
In addition, the optimum position of the conformal bent cigar function can considerably influence both CMA-ES and PSO. When the optimum position is set around the thinner area, CMA-ES and PSO are hardly able to find it with the preset number of fitness evaluations. We believe these results generalize to other ill conditional problems and other evolutionary algorithms.

\par
One of the most important implications from the results could be the appropriate use of restart strategy.
In general, the current restart mechanism in CMA-ES is activated only when the search agent gets stuck somewhere or the fitness value has no improvement for some time. However, it does not work well in the conformal bent cigar function since the fitness value is continuously decreasing and population mean is also slowly moving. Furthermore, the performance variation on the same function indicates the starting search point has considerable influence on optimization of the function. In this case, it could be a better choice to restart the search rather than follow one way to the end. But then again, it is a tradeoff between keeping the current promising but slow trend and restarting a possible faster trial.

\par
Another consideration can be adopting more intelligent strategy to aid the optimizer in recognizing the general environment. Obviously, the winding valley is ring-shaped and has clear regularity. We can deduce that a certain additional machine learning could be helpful to sufficiently utilize the previous search experience.And it may provide inspiration and help on reinforcement learning issues.

\section*{Acknowledgments}

This work was supported by National Natural Science Foundation of China under Grant No. 61872419, No. 61573166, No. 61572230. Shandong Provincial Key R\&D Program under Grant No. 2019GGX101041, No. 2018CXGC0706. Taishan Scholars Program of Shandong Province, China, under Grant No. tsqn201812077.

\bibliographystyle{IEEEtran}
\bibliography{bib/ref}

\end{document}